\documentclass{article}
\usepackage[T1]{fontenc}
\usepackage{spconf,amsmath,graphicx}
\usepackage{amsmath,graphicx,booktabs,amssymb}
\usepackage{multirow}
\usepackage{algorithm}
\usepackage[noend]{algpseudocode}
\usepackage{cite}
\usepackage{url}
\usepackage{xcolor}
\usepackage[hidelinks]{hyperref}

\title{LOGIC: Efficient and Robust Contextual Biasing for Speech LLMs via Logit-Space Integration}
\name{Peidong Wang, Jian Xue, Jinyu Li}
\address{Microsoft, USA}

\begin{document}
\ninept
\maketitle

\begin{abstract}
Recognizing entity phrases remains a critical challenge for speech large language models. Existing prompting methods lack an explicit decoding-time biasing weight, limiting their controllability. Generative error correction methods can introduce hallucinated over-corrections. To address these limitations, we propose LOGIC (logit-space integration for contextual biasing), a robust framework operating directly in the logit space. By decoupling context injection from input processing, LOGIC enables explicit control over the biasing strength. Extensive experiments with an open-source speech large language model across 11 locales demonstrate that LOGIC achieves an average 9\% relative reduction in entity word error rate, with an average false alarm rate increase of 0.3\% and a 2.8\% relative runtime overhead. When combined with prompting, LOGIC can reduce entity word error rate by 5\% relative to the prompt-only method.
\end{abstract}


\section{Introduction}
\label{sec:intro}

Recent advances in large language models (LLMs) have revolutionized the landscape of artificial intelligence. By integrating speech modalities directly into the modeling process, speech LLMs enable users to interact with intelligent agents using natural voice \cite{hurst2024gpt, team2023gemini, abouelenin2025phi}. Unlike traditional cascaded systems using automatic speech recognition (ASR) \cite{graves2012asr,wang2018utterance,wang2021mute,wang2022lamassu} followed by an LLM \cite{elnokrashy2022language}, speech LLMs perform end-to-end modeling, effectively capturing paralinguistic cues and semantic nuances. However, despite their impressive conversational capabilities, these models often struggle with the accurate recognition of rare or domain-specific entities such as contact names, technical jargon, or obscure geographic locations that rarely appear in general training corpora \cite{gaido2022we,modrzejewski2020incorporating}. Accurate recognition of these entity phrases is critical for user satisfaction in personalized applications \cite{gong2024contextual,wang2026phrased}.

A straightforward approach to address this limitation is prompting \cite{fathullah2024prompting,hu2025chain}. With appropriate instructions, speech LLMs can be directed to bias towards specific entities provided in the text prompt. For instance, a user might provide a list of contact names in the system prompt. However, this method faces limitations in scalability and efficiency. Prompting with long entity lists can increase latency and inference costs. Moreover, it does not directly expose a numerical biasing weight for refined biasing control.

Another commonly adopted method for entity recognition is {generative error correction (GEC)} \cite{yang2023generative,ghosh2024failing}, which uses a separate text-based LLM to rewrite ASR outputs based on a bias list. While GEC allows for correcting phonetic errors, it can generate entity phrases that are not even spoken. This significantly degrades user experiences.

In this paper, we propose {LOGIC} ({LOG}it-space {I}ntegration for {C}ontextual biasing), an efficient and robust framework for LLMs. LOGIC complements prompting by injecting bias information directly into the decoding process via a phrase-matching mechanism \cite{pundak2018deep,le2021contextualized,sainath2023improving,wang2024contextual,bhattacharjee2024contextual}.
We validate LOGIC on large-scale person name (PNAME) test sets across {11 distinct locales}. Results show that our method achieves an average 9\% relative improvement in entity word error rate (EWER) with an average false alarm rate (FAR) increase of 0.3\%. When combined with prompting, LOGIC provides an additional 5\% relative EWER reduction over the prompt-only method. Additionally, we conduct efficiency profiling, demonstrating that our sparse vocabulary implementation introduces minimal latency overhead of 2.8\% in real-time factor (RTF), making LOGIC a practical solution for large-scale deployment.

\section{Related Work}
\label{sec:related}

\subsection{Contextual Biasing in Traditional ASR}
Contextual biasing has been extensively studied in hybrid and end-to-end (E2E) ASR systems. The most common approach is {shallow fusion} \cite{zhao2019shallow}, where an external context language model (CLM) is interpolated with the ASR model scores during beam search. While effective, shallow fusion often struggles with phrases that are acoustically similar to the ground truth. More advanced methods like {deep fusion} \cite{modrzejewski2020incorporating} integrate context embeddings directly into the encoder or decoder layers. However, these methods require retraining or fine-tuning of the underlying model, which can be impractical for large-scale LLMs.

\subsection{LLM Adaptation: Prompting and Fine-tuning}
For LLMs, contextual customization is typically achieved via {prompting} \cite{liu2023pre} or {fine-tuning} (e.g., LoRA) \cite{hu2022lora}. As discussed in Section \ref{sec:intro}, prompting increases the input length and cannot easily control the biasing weight. In addition, fine-tuning requires additional computation to adapt the underlying model.

\subsection{Generative Error Correction (GEC)}
GEC leverages the text-generation capabilities of LLMs to correct ASR transcripts \cite{yang2023generative,ma2023can,du2024cot,ghosh2024failing}. A GEC model takes the N-best hypotheses and a list of entities as input to generate a revised transcript.
However, a critical flaw in GEC is the hallucination issue. Due to the complete rewriting of the transcriptions, some entity phrases that are not spoken can be mistakenly added to the output.
This introduction of non-existent entities results in a bad user experience.

\section{Proposed Method: LOGIC}
\label{sec:method}

In this section, we present the architecture of LOGIC. We first describe the phrase representation and logit integration, followed by our robust scoring strategies and efficiency optimizations.

\subsection{Logit Integration}
\label{ssec:trie}
Let $\mathbf{O} = \{E_1, \dots, E_N\}$ be the set of entity phrases.
A critical challenge in biasing LLMs, compared to traditional phoneme-based ASR, is the complexity of subword tokenization. An entity string may be tokenized differently depending on its surrounding context. For example, the name ``Alex'' might be tokenized as `[\_Alex]' (with leading space) after whitespace, but as `[Alex]' or `[Al, ex]' in other contexts.
To handle this, we employ a {best-of-K tokenization} strategy. For each entity, we generate multiple candidate token sequences (e.g., with and without leading whitespace) and select the shortest valid tokenization as the canonical representation. Additionally, we build a {token equivalence} mapping that identifies tokens differing only in whitespace conventions (e.g., `\_the' $\leftrightarrow$ `the'). This supports matching across these tokenization boundary variations.

All entity phrases are stored in a {phrase tensor} $\mathbf{P} \in \mathbb{Z}^{N \times L_{\max}}$, where $N$ is the number of phrases and $L_{\max}$ is the maximum phrase length (shorter phrases are padded). For each phrase $i$, we maintain a {matching state} $\sigma_t^{(b,i)} \in \{0, 1, \dots, L_i\}$ that tracks how many consecutive tokens of phrase $i$ have been matched by hypothesis $b$ up to time step $t$. The state transitions use a first-token restart rule: if a token matches the expected token at position $\sigma_t^{(b,i)}$ of an incomplete phrase $i$ (including via the token equivalence map), the state advances to $\sigma_t^{(b,i)} + 1$. On mismatch or after a complete match, LOGIC checks whether the token matches the {first} token of phrase $i$; if so, the state restarts at 1 (partial match restart), otherwise it resets to 0.

We define the {matching potential} of a hypothesis as the maximum matching depth across all phrases:
\begin{equation}
    \Pi_t = \max_{i \in \{1,\dots,N\}} \sigma_t^{(b,i)}
\end{equation}

During auto-regressive decoding, at time step $t$, let $\mathbf{z}_t \in \mathbb{R}^{|\mathcal{V}|}$ be the logits output by the LLM. For each candidate token $k$ in the vocabulary, we compute the hypothetical next states $\hat{\sigma}_{t+1}^{(b,i)}(k)$ for all phrases $i$, and the resulting potential $\hat{\Pi}_{t+1}(k)$. Before mismatch rectification, the biased logits are computed as:
\begin{equation}
    \mathbf{z}'_t[k] = \mathbf{z}_t[k] + \lambda \cdot \underbrace{\left(\hat{\Pi}_{t+1}(k) - \Pi_t\right)}_{\Delta\Pi_k}
    \label{eq:biasing}
\end{equation}
where $\lambda$ is the per-token bonus weight and $\Delta\Pi_k$ is the potential difference. Tokens that increase matching potential by one ($\Delta\Pi_k = 1$) receive a bonus $+\lambda$, tokens that maintain progress ($\Delta\Pi_k = 0$) are unaffected, and tokens that reduce matching potential ($\Delta\Pi_k < 0$) may receive a penalty (see Section~\ref{ssec:strategies}). With RSR enabled, negative increments are replaced by $-\lambda\Pi_t$; otherwise they are clipped to zero. This {potential-based scoring} rewards tokens proportionally to how much they advance matching progress.

\begin{algorithm}[t]
\caption{LOGIC Inference Step (per hypothesis)}
\label{alg:logic}
\begin{algorithmic}[1]
\State {Input:} Logits $\mathbf{z}_t$, States $\{\sigma_t^{(i)}\}_{i=1}^N$, Phrases $\mathbf{P}$
\State {Params:} $\lambda$ (Bonus), IPB (True), RSR (True)
\State {Output:} Modified $\mathbf{z}'_t$, New States $\{\sigma_{t+1}^{(i)}\}$
\State Compute current potential: $\Pi_t \leftarrow \max_i \sigma_t^{(i)}$
\State $K_{\text{sparse}} \leftarrow$ phrase tokens and supported equivalents
\State $\mathbf{z}'_t \leftarrow \mathbf{z}_t$
\If{RSR is enabled}
    \State $\mathbf{z}'_t \leftarrow \mathbf{z}_t - \lambda\Pi_t$
\EndIf
\For{each token $k \in K_{\text{sparse}}$}
    \For{each phrase $i$}
        \State Compute $\hat{\sigma}_{t+1}^{(i)}(k)$ via first-token restart matching
    \EndFor
    \State $\hat{\Pi}_{t+1}(k) \leftarrow \max_i \hat{\sigma}_{t+1}^{(i)}(k)$
    \State $\Delta\Pi_k \leftarrow \hat{\Pi}_{t+1}(k) - \Pi_t$
    \State $\mathbf{z}'_t[k] \leftarrow \mathbf{z}_t[k] + \lambda\max(\Delta\Pi_k,0)$
\EndFor
\If{RSR is enabled}
    \For{$k \in K_{\text{sparse}}$ where $\Delta\Pi_k < 0$}
        \State $\mathbf{z}'_t[k] \leftarrow \mathbf{z}_t[k] - \lambda\Pi_t$
    \EndFor
\EndIf
\State $y_t \leftarrow \text{Decode}(\mathbf{z}'_t)$
\For{each phrase $i$}
    \If{$\sigma_t^{(i)} < L_i$ and $y_t$ matches the expected token}
        \State $\sigma_{t+1}^{(i)} \leftarrow \sigma_t^{(i)} + 1$
    \ElsIf{$y_t$ matches first token of phrase $i$}
        \State $\sigma_{t+1}^{(i)} \leftarrow 1$
    \Else
        \State $\sigma_{t+1}^{(i)} \leftarrow 0$
    \EndIf
\EndFor
\State \Return $\mathbf{z}'_t, \{\sigma_{t+1}^{(i)}\}$
\end{algorithmic}
\end{algorithm}

\subsection{Robust Inference Strategies}
\label{ssec:strategies}

\subsubsection{Immediate Prefix Boosting (IPB)}

For speech LLMs with large vocabularies, the first token is often semantically significant, e.g., a Chinese surname. Discarding its bias can degrade recall for short entities.

In LOGIC, we adopt {immediate prefix boosting}. The full bonus $\lambda$ is applied from the first token when no phrase is active, including transitions from state $\sigma = 0 \to 1$:
\begin{equation}
    \text{Bias}(k) = \lambda \cdot \Delta\Pi_k, \quad \Delta\Pi_k \geq 0
\end{equation}
This increases sensitivity to entity onsets. While this aggressive strategy can increase false alarm risks, we combine it with the following rectification mechanism.

\subsubsection{Retroactive Score Rectification (RSR)}
Generative models can exhibit a ``sunk cost'' effect: if a hypothesis accumulates bonuses for a partial match (e.g., matching the prefix ``Alex'' when the audio is ``Algorithm''), the inflated score may cause the model to hallucinate the remainder of the entity (``Alexander'').

To control partial-match scores, we introduce {retroactive score rectification}. When a candidate token $k$ would cause a decrease in matching potential ($\Delta\Pi_k < 0$), we apply a penalty proportional to the current matching depth:
\begin{equation}
    \text{Penalty}(k) = -\lambda \cdot \Pi_t, \quad \text{if } \Delta\Pi_k < 0
\end{equation}
This penalty replaces the negative increment in Eq.~\eqref{eq:biasing}. RSR adjusts the current logits rather than revising emitted tokens or exactly canceling normalized sequence probabilities. Together, IPB and RSR provide a balance between entity recall and FAR, evaluated jointly in our experiments.

\subsection{System Implementation}
\label{ssec:efficiency_impl}
Efficiency is paramount for real-time interaction. We implement LOGIC on the {vLLM} platform as a custom \texttt{LogitsProcessor} class.
\begin{itemize}
    \item {Vectorized Computation}: Phrase matching and score computation are vectorized using PyTorch tensor operations on GPU, processing phrases and candidate tokens without Python-level loops in the scoring path.
    \item {Sparse Vocabulary}: A naive iteration over the full vocabulary $|\mathcal{V}|$ (often $>150$k) is computationally expensive. Since only a small subset of tokens appear in any phrase, we pre-compute a sparse vocabulary set $K_{\text{sparse}} \subset \mathcal{V}$ containing the unique phrase tokens and supported equivalents (typically $|K_{\text{sparse}}| \approx 100$--$1000$). Phrase-dependent scores are computed only for $K_{\text{sparse}}$ and scattered back to the full vocabulary, reducing the candidate set by two to three orders of magnitude. A shared background penalty handles mismatches outside this set when RSR is enabled.
    \item {Pre-Computed Caches}: Phrase length tensors, first-token vectors, phrase index tensors, and token equivalence lookup tables are all pre-computed during phrase setup and cached for reuse across decoding steps, avoiding redundant computation in the hot path.
\end{itemize}

\section{Experimental Setup}
\label{sec:exp}

\subsection{Datasets and Configuration}
We conduct our experiments using the open-source {Phi-4-mini} model \cite{abouelenin2025phi}. The ASR evaluation is performed on an {internally curated ASR test set} covering {11 locales}: en-US, ko-KR, zh-CN, es-ES, es-MX, en-GB, de-DE, fr-FR, it-IT, ja-JP, and pt-BR.

The test sets specifically focus on person names (PNAMEs), representing a challenging problem where rare entities are embedded in natural speech. The volume of the test entities varies by locale, ranging from approximately 460 entities (de-DE) to over 2100 entities (it-IT) per locale.

\subsection{Metrics}
We report standard word error rate (WER) and specific entity metrics:
\begin{itemize}
    \item {EWER}: Error rate calculated specifically on the target entity regions.
    \item {Entity Recall}: The percentage of ground-truth entities that are correctly recognized.
    \item {FAR}: The rate of incorrect entity insertions per utterance.
    \item {RTF}: Processing time divided by audio duration, to measure efficiency.
\end{itemize}
The average relative EWER improvements in Tables~\ref{tab:main_results} and~\ref{tab:aggressive_full} are unweighted averages across locales.

\section{Results and Analysis}
\label{sec:results}
We conducted experiments in two primary configurations: a {robust} setting ($\lambda=0.4$, with lower FAR) and an {aggressive} setting ($\lambda=1.0$, targeting higher recall).

\begin{table}[htb]
\centering
\caption{Main Results: LOGIC under {Robust} setting. IPB and RSR are both enabled. {Rel. Imp.} denotes the relative improvement of EWER. This setting prioritizes low False Alarm Rate. $\Delta$FAR is in percentage points.}
\label{tab:main_results}
\resizebox{\linewidth}{!}{%
\begin{tabular}{l|c|ccc|cc}
\toprule
{Locale} & {System} & {WER} & {EWER} & {Recall} & {Rel. Imp.} & {$\Delta$ FAR} \\
\midrule
\multirow{2}{*}{en-US} & Baseline & 9.8 & 22.36 & 75.00 & \multirow{2}{*}{11\%} & \multirow{2}{*}{0.25} \\
 & LOGIC & 9.7 & {19.83} & 77.53 & & \\ \midrule
\multirow{2}{*}{ko-KR} & Baseline & 7.6 & 16.77 & 65.75 & \multirow{2}{*}{6\%} & \multirow{2}{*}{0.00} \\
 & LOGIC & 7.6 & {15.83} & 68.04 & & \\ \midrule
\multirow{2}{*}{zh-CN} & Baseline & 7.3 & 28.51 & 59.02 & \multirow{2}{*}{9\%} & \multirow{2}{*}{0.94} \\
 & LOGIC & 7.3 & {25.88} & 63.47 & & \\ \midrule
\multirow{2}{*}{es-ES} & Baseline & 7.9 & 12.93 & 84.01 & \multirow{2}{*}{5\%} & \multirow{2}{*}{0.31} \\
 & LOGIC & 7.9 & {12.28} & 84.64 & & \\ \midrule
\multirow{2}{*}{es-MX} & Baseline & 7.0 & 14.38 & 80.00 & \multirow{2}{*}{17\%} & \multirow{2}{*}{0.00} \\
 & LOGIC & 7.0 & {11.87} & 83.71 & & \\ \midrule
\multirow{2}{*}{en-GB} & Baseline & 8.3 & 23.65 & 74.58 & \multirow{2}{*}{5\%} & \multirow{2}{*}{0.18} \\
 & LOGIC & 8.2 & {22.57} & 75.70 & & \\ \midrule
\multirow{2}{*}{de-DE} & Baseline & 6.9 & 21.66 & 75.92 & \multirow{2}{*}{14\%} & \multirow{2}{*}{0.00} \\
 & LOGIC & 6.8 & {18.72} & 79.60 & & \\ \midrule
\multirow{2}{*}{fr-FR} & Baseline & 6.9 & 13.50 & 83.96 & \multirow{2}{*}{19\%} & \multirow{2}{*}{0.47} \\
 & LOGIC & 6.9 & {10.95} & 87.26 & & \\ \midrule
\multirow{2}{*}{it-IT} & Baseline & 7.7 & 14.67 & 82.67 & \multirow{2}{*}{2\%} & \multirow{2}{*}{0.36} \\
 & LOGIC & 7.8 & {14.37} & 82.67 & & \\ \midrule
\multirow{2}{*}{ja-JP} & Baseline & 9.4 & 42.10 & 50.63 & \multirow{2}{*}{6\%} & \multirow{2}{*}{0.00} \\
 & LOGIC & 9.5 & {39.51} & 53.97 & & \\ \midrule
\multirow{2}{*}{pt-BR} & Baseline & 12.0 & 30.21 & 60.22 & \multirow{2}{*}{6\%} & \multirow{2}{*}{0.82} \\
 & LOGIC & 12.0 & {28.46} & 62.13 & & \\ \midrule
\multicolumn{2}{c|}{{Average}} & - & - & - & {9\%} & {0.30} \\
\bottomrule
\end{tabular}%
}
\end{table}

\subsection{Results in Robust Setting}
Table \ref{tab:main_results} summarizes the performance under the robust configuration. Although named entities constitute only a small fraction of each utterance, correcting them can affect generic WER. Across all 11 locales, LOGIC yields comparable generic WERs to the unbiased baseline, with changes of at most 0.1 percentage points at the reported precision, including slight increases in it-IT and ja-JP.

For entity results, LOGIC achieves an average {9\% relative reduction in EWER}, with gains observed consistently across all locales. The improvements are especially substantial in Romance languages such as fr-FR (19\%) and es-MX (17\%), and in English (en-US, 11\%). Notably, even in zh-CN and ja-JP, where tokenization boundaries can be ambiguous, LOGIC still delivers 9\% and 6\% improvements respectively, demonstrating its applicability across these locales.

Perhaps most importantly, these recall gains come with a small average increase in false alarms. Despite the aggressive first-token boosting of IPB, the average FAR increase is {0.3\%}, with several locales (ko-KR, es-MX) showing zero additional false alarms at the reported precision.

\subsection{Comparison with Prompt-Only}
\label{ssec:prompt}
We further compare LOGIC combined with prompting against the prompt-only method on the same PNAME evaluation set and similar experimental settings. As shown in Table~\ref{tab:prompt_comparison}, adding LOGIC provides an additional 5\% relative reduction in EWER over prompt-only. This result demonstrates that LOGIC can complement prompting.

\begin{table}[htb]
\centering
\caption{Comparison with prompt-only on the same PNAME evaluation set, using similar experimental settings. The EWER reduction is relative to the prompt-only method.}
\label{tab:prompt_comparison}
\small
\setlength{\tabcolsep}{4pt}
\begin{tabular}{llr}
\toprule
Baseline & Evaluated system & Rel. EWER reduction\\
\midrule
Prompt-only & Prompt+LOGIC & {5\%}\\
\bottomrule
\end{tabular}
\end{table}

\subsection{Sensitivity and Potential Analysis}
To investigate the sensitivity of LOGIC's capability, we conducted a sensitivity study by increasing the biasing bonus $\lambda$ from 0.4 to 1.0. This {Aggressive} setting targets applications where recall is paramount (e.g., searching for a specific contact). Note that we use a unified biasing bonus for different locales in the table. The results can be further improved if we use specified biasing bonuses for different locales.

\begin{table}[h]
\centering
\caption{Sensitivity Analysis: Relative EWER Improvement (Robust vs. Aggressive) across all 11 Locales. Gain is the difference in relative improvements, in percentage points.}
\label{tab:aggressive_full}
\resizebox{0.7\linewidth}{!}{%
\begin{tabular}{lccc}
\toprule
\textbf{Locale} & \textbf{Robust} & \textbf{Aggressive} & \textbf{Gain} \\
\midrule
en-US & 11\% & 18\% & +7 \\
zh-CN & 9\% & 20\% & +11 \\
fr-FR & 19\% & 29\% & +10 \\
es-MX & 17\% & 28\% & +11 \\
de-DE & 14\% & 24\% & +10 \\
pt-BR & 6\% & 21\% & +15 \\
en-GB & 5\% & 12\% & +7 \\
ko-KR & 6\% & 10\% & +4 \\
es-ES & 5\% & 11\% & +6 \\
ja-JP & 6\% & 10\% & +4 \\
it-IT & 2\% & 5\% & +3 \\
\midrule
\textbf{Average} & \textbf{9\%} & \textbf{17\%} & \textbf{+8} \\
\bottomrule
\end{tabular}
}
\end{table}

As shown in Table \ref{tab:aggressive_full}, this aggressive setting unlocks significant potential, boosting the average relative improvement to {17\%}. Specifically, locales such as fr-FR and es-MX see improvements jumping to {29\%} and {28\%} respectively. Even in zh-CN, the improvement more than doubles from 9\% to 20\%.
This demonstrates that LOGIC is highly responsive to tuning. The LOGIC framework provides a controllable lever: users can choose a lower $\lambda$ for open-ended dictation or a higher $\lambda$ for constrained command-and-control tasks. 

\subsection{Qualitative Analysis}
\label{ssec:qualitative}
To demonstrate the effectiveness of LOGIC in recovering named entities, we compare its output against the baseline (no biasing) in Table \ref{tab:logic_correction}.

\begin{table}[htb]
\centering
\caption{Qualitative Comparison: Baseline Errors vs. LOGIC Corrections. LOGIC successfully recovers entities in these examples.}
\label{tab:logic_correction}
\resizebox{0.9\linewidth}{!}{%
\begin{tabular}{p{0.42\linewidth}|p{0.42\linewidth}}
\toprule \textbf{Baseline (Error)} & \textbf{LOGIC (Correct)} \\
\midrule
...that \textbf{mirror} referred to. & ...that \textbf{Mira} referred to. \\
\textit{(Phonetic confusion)} & \textit{(Bias on `Mira' overcomes acoustic ambiguity)} \\
\midrule
\textbf{If I} who are we waiting for? & \textbf{Yvonne}, who are we waiting for? \\
 \textit{(Segmentation error at onset)} & \textit{(Entity onset recovered)} \\
\bottomrule
\end{tabular}
}
\end{table}

\begin{table}[!htb]
\centering
\caption{Efficiency Analysis (Real-Time Factor 90th percentile). ``LOGIC'' uses the sparse vocabulary optimization. The number of phrases in the phrase list is 2210.}
\label{tab:efficiency}
\resizebox{0.9\linewidth}{!}{%
\begin{tabular}{lcc}
\toprule
{System} & {RTF} & {Overhead} \\
\midrule
Baseline (No Biasing) & 0.0990 & - \\
LOGIC (Sparse Vocabulary) & 0.1018 & +2.8\% \\
\bottomrule
\end{tabular}
}
\end{table}

Table \ref{tab:logic_correction} illustrates two error patterns. First, for phonetic ambiguities like ``Mira/mirror,'' LOGIC uses the potential-based scoring to boost the entity path without breaking the generation. Second, for segmentation errors like ``Yvonne/If I,'' the {Immediate Prefix Boosting} (IPB) can help the rare name compete with common function words at the entity onset.

\subsection{Computational Efficiency}
A common skepticism of contextual biasing for LLMs is the latency overhead. We benchmarked the real-time factor (RTF) of our proposed system against the baseline on a standard inference server using NVIDIA A100 GPUs.
As shown in Table \ref{tab:efficiency}, the baseline system operates at an RTF of 0.099. Enabling LOGIC increases the RTF to only 0.1018. This represents a marginal overhead of approximately {2.8\%}.
This efficiency is mainly attributed to our sparse vocabulary optimization. Instead of iterating over the full vocabulary size ($>150,000$ tokens), our implementation only processes the unique tokens that appear in any phrase ($\approx 100$--$1,000$ tokens typically), computes biasing scores for this sparse set, and scatters the results back to the full logit tensor. 

\section{Conclusion}
\label{sec:conc}
We presented LOGIC, a contextual biasing framework that operates directly in the logit space of speech LLMs. By representing entity phrases as a compact phrase tensor and scoring candidate tokens through a potential-based mechanism, LOGIC provides soft guidance without modifying the underlying model. Two complementary strategies, IPB for boosting entity recall from the very first token, and RSR for penalizing mismatches, provide more controllability. A sparse vocabulary optimization method keeps the computational overhead small.
Our evaluation across 11 locales on the Phi-4-mini model demonstrates that LOGIC achieves a 9\% relative EWER improvement with only a 0.3\% increase in FAR and a 2.8\% additional computation overhead. Adding LOGIC to prompting provides an additional 5\% relative EWER reduction over prompt-only on the same evaluation set.
In future work, we plan to explore: (1) {semantic biasing}, integrating embedding-based retrieval to support fuzzy matching beyond strict token-level constraints; (2) {multi-entity co-occurrence modeling}, leveraging inter-entity correlations to improve disambiguation; and (3) {end-to-end training}, designing loss functions that simulate LOGIC's inference dynamics during training to further improve discriminability.
\label{lastmainpage}

\clearpage
\bibliographystyle{IEEEbib}
\bibliography{refs}

@article{le2021contextualized,
  title={Contextualized streaming end-to-end speech recognition with trie-based deep biasing and shallow fusion},
  author={Le, Duc and Jain, Mahaveer and Keren, Gil and Kim, Suyoun and Shi, Yangyang and Mahadeokar, Jay and Chan, Julian and Shangguan, Yuan and Fuegen, Christian and Kalinli, Ozlem and others},
  journal={arXiv preprint arXiv:2104.02194},
  year={2021}
}

@inproceedings{bhattacharjee2024contextual,
  title={Contextual Biasing Methods for Improving Rare Word Detection in Automatic Speech Recognition},
  author={Bhattacharjee, Mrinmoy and Nigmatulina, Iuliia and Prasad, Amrutha and Rangappa, Pradeep and Madikeri, Srikanth and Motlicek, Petr and Helmke, Hartmut and Kleinert, Matthias},
  booktitle={ICASSP 2024-2024 IEEE International Conference on Acoustics, Speech and Signal Processing (ICASSP)},
  pages={12652--12656},
  year={2024},
  organization={IEEE}
}

@inproceedings{sainath2023improving,
  title={Improving contextual biasing with text injection},
  author={Sainath, Tara N and Prabhavalkar, Rohit and Caseiro, Diamantino and Rondon, Pat and Allauzen, Cyril},
  booktitle={ICASSP 2023-2023 IEEE International Conference on Acoustics, Speech and Signal Processing (ICASSP)},
  pages={1--5},
  year={2023},
  organization={IEEE}
}

@inproceedings{pundak2018deep,
  title={Deep context: end-to-end contextual speech recognition},
  author={Pundak, Golan and Sainath, Tara N and Prabhavalkar, Rohit and Kannan, Anjuli and Zhao, Ding},
  booktitle={2018 IEEE spoken language technology workshop (SLT)},
  pages={418--425},
  year={2018},
  organization={IEEE}
}

@inproceedings{fathullah2024prompting,
  title={Prompting large language models with speech recognition abilities},
  author={Fathullah, Yassir and Wu, Chunyang and Lakomkin, Egor and Jia, Junteng and Shangguan, Yuan and Li, Ke and Guo, Jinxi and Xiong, Wenhan and Mahadeokar, Jay and Kalinli, Ozlem and others},
  booktitle={ICASSP 2024-2024 IEEE International Conference on Acoustics, Speech and Signal Processing (ICASSP)},
  pages={13351--13355},
  year={2024},
  organization={IEEE}
}

@article{gong2024contextual,
  title={Contextual biasing speech recognition in speech-enhanced large language model},
  author={Gong, Xun and Lv, Anqi and Wang, Zhiming and Qian, Yanmin},
  journal={Proc. Interspeech. ISCA},
  pages={257--261},
  year={2024}
}

@article{hu2022lora,
  title={Lora: Low-rank adaptation of large language models.},
  author={Hu, Edward J and Shen, Yelong and Wallis, Phillip and Allen-Zhu, Zeyuan and Li, Yuanzhi and Wang, Shean and Wang, Lu and Chen, Weizhu and others},
  journal={ICLR},
  volume={1},
  number={2},
  pages={3},
  year={2022}
}

@article{liu2023pre,
  title={Pre-train, prompt, and predict: A systematic survey of prompting methods in natural language processing},
  author={Liu, Pengfei and Yuan, Weizhe and Fu, Jinlan and Jiang, Zhengbao and Hayashi, Hiroaki and Neubig, Graham},
  journal={ACM computing surveys},
  volume={55},
  number={9},
  pages={1--35},
  year={2023},
  publisher={ACM New York, NY}
}

@inproceedings{zhao2019shallow,
  title={Shallow-Fusion End-to-End Contextual Biasing.},
  author={Zhao, Ding and Sainath, Tara N and Rybach, David and Rondon, Pat and Bhatia, Deepti and Li, Bo and Pang, Ruoming},
  booktitle={Interspeech},
  pages={1418--1422},
  year={2019}
}

@article{ma2023can,
  title={Can generative large language models perform asr error correction?},
  author={Ma, Rao and Qian, Mengjie and Manakul, Potsawee and Gales, Mark and Knill, Kate},
  journal={arXiv preprint arXiv:2307.04172},
  year={2023}
}

@inproceedings{yang2023generative,
  title={Generative speech recognition error correction with large language models and task-activating prompting},
  author={Yang, Chao-Han Huck and Gu, Yile and Liu, Yi-Chieh and Ghosh, Shalini and Bulyko, Ivan and Stolcke, Andreas},
  booktitle={2023 IEEE Automatic Speech Recognition and Understanding Workshop (ASRU)},
  pages={1--8},
  year={2023},
  organization={IEEE}
}

@article{ghosh2024failing,
  title={Failing Forward: Improving Generative Error Correction for ASR with Synthetic Data and Retrieval Augmentation},
  author={Ghosh, Sreyan and Rasooli, Mohammad Sadegh and Levit, Michael and Wang, Peidong and Xue, Jian and Manocha, Dinesh and Li, Jinyu},
  journal={arXiv preprint arXiv:2410.13198},
  year={2024}
}

@inproceedings{wang2026phrased,
  title={PHRASED: Phrase Dictionary Biasing for Speech Translation},
  author={Wang, Peidong and Xue, Jian and Zhao, Rui and Chen, Junkun and Subramanian, Aswin and Li, Jinyu},
  booktitle={Proc. of ICASSP},
  pages={1--5},
  year={2026},
  organization={IEEE}
}

@inproceedings{hu2025chain,
  title={Chain-of-thought prompting for speech translation},
  author={Hu, Ke and Chen, Zhehuai and Yang, Chao-Han Huck and {\.Z}elasko, Piotr and Hrinchuk, Oleksii and Lavrukhin, Vitaly and Balam, Jagadeesh and Ginsburg, Boris},
  booktitle={Proc. of ICASSP},
  pages={1--5},
  year={2025},
  organization={IEEE}
}

@article{du2024cot,
  title={CoT-ST: Enhancing LLM-based Speech Translation with Multimodal Chain-of-Thought},
  author={Du, Yexing and Ma, Ziyang and Yang, Yifan and Deng, Keqi and Chen, Xie and Yang, Bo and Xiang, Yang and Liu, Ming and Qin, Bing},
  journal={arXiv preprint arXiv:2409.19510},
  year={2024}
}

@article{abouelenin2025phi,
  title={Phi-4-Mini Technical Report: Compact yet Powerful Multimodal Language Models via Mixture-of-LoRAs},
  author={Abouelenin, Abdelrahman and Ashfaq, Atabak and Atkinson, Adam and Awadalla, Hany and Bach, Nguyen and Bao, Jianmin and Benhaim, Alon and Cai, Martin and Chaudhary, Vishrav and Chen, Congcong and others},
  journal={arXiv preprint arXiv:2503.01743},
  year={2025}
}

@article{team2023gemini,
  title={Gemini: a family of highly capable multimodal models},
  author={Team, Gemini and Anil, Rohan and Borgeaud, Sebastian and Alayrac, Jean-Baptiste and Yu, Jiahui and Soricut, Radu and Schalkwyk, Johan and Dai, Andrew M and Hauth, Anja and Millican, Katie and others},
  journal={arXiv preprint arXiv:2312.11805},
  year={2023}
}

@article{hurst2024gpt,
  title={Gpt-4o system card},
  author={Hurst, Aaron and Lerer, Adam and Goucher, Adam P and Perelman, Adam and Ramesh, Aditya and Clark, Aidan and Ostrow, AJ and Welihinda, Akila and Hayes, Alan and Radford, Alec and others},
  journal={arXiv preprint arXiv:2410.21276},
  year={2024}
}

@inproceedings{modrzejewski2020incorporating,
  title={Incorporating external annotation to improve named entity translation in NMT},
  author={Modrzejewski, Maciej and Exel, Miriam and Buschbeck, Bianka and Ha, Thanh-Le and Waibel, Alex},
  booktitle={Proceedings of the 22nd annual conference of the european association for machine translation},
  pages={45--51},
  year={2020}
}

@inproceedings{wang2022lamassu,
  title={{LAMASSU}: Streaming Language-Agnostic Multilingual Speech Recognition and Translation Using Neural Transducers},
  author={Wang, Peidong and Sun, Eric and Xue, Jian and Wu, Yu and Zhou, Long and Gaur, Yashesh and Liu, Shujie and Li, Jinyu},
  booktitle={Proc. Interspeech},
  pages={57--61},
  year={2023},
}

@inproceedings{wang2018utterance,
  title={Utterance-wise recurrent dropout and iterative speaker adaptation for robust monaural speech recognition},
  author={Wang, P. and Wang, D. L.},
  booktitle={Proc. of ICASSP},
  pages={4814--4818},
  year={2018},
  organization={IEEE}
}

@inproceedings{wang2021mute,
  title={Multitask training with text data for end-to-end speech recognition},
  author={Wang, P. and Sainath, T. N. and Weiss, R. J.},
  booktitle={Proc. of INTERSPEECH},
  pages={2566--2570},
  year={2021}
}

@article{elnokrashy2022language,
  title={Language Tokens: A Frustratingly Simple Approach Improves Zero-Shot Performance of Multilingual Translation},
  author={ElNokrashy, M. and Hendy, A. and Maher, M. and Afify, M. and Awadalla, H. H.},
  journal={arXiv preprint arXiv:2208.05852},
  year={2022}
}

@article{graves2012asr,
  title={Sequence transduction with recurrent neural networks},
  author={Graves, A.},
  journal={arXiv preprint arXiv:1211.3711},
  year={2012}
}

@inproceedings{gaido2022we,
  title={Who Are We Talking About? Handling Person Names in Speech Translation},
  author={Gaido, Marco and Negri, Matteo and Turchi, Marco},
  booktitle={Proc. IWSLT},
  pages={62--73},
  year={2022}
}

@inproceedings{wang2024contextual,
  title={Contextual Biasing with the Knuth-Morris-Pratt Matching Algorithm},
  author={Wang, Weiran and Wu, Zelin and Caseiro, Diamantino and Munkhdalai, Tsendsuren and Sim, Khe Chai and Rondon, Pat and Pundak, Golan and Song, Gan and Prabhavalkar, Rohit and Meng, Zhong and others},
  booktitle={Proc. Interspeech 2024},
  pages={282--286},
  year={2024}
}
\end{document}